\documentclass[10pt,twocolumn,letterpaper]{article}

\usepackage[pagenumbers]{cvpr} % To force page numbers, e.g. for an arXiv version

\usepackage[dvipsnames]{xcolor}

\usepackage{pifont}
\newcommand{\cmark}{\ding{51}}%
\newcommand{\xmark}{\ding{55}}%

\usepackage{colortbl}
\definecolor{lightgray}{gray}{0.95} % 90% white
\usepackage{xcolor}
\usepackage{arydshln}
\usepackage{makecell}

\definecolor{cvprblue}{rgb}{0.21,0.49,0.74}
\usepackage[pagebackref,breaklinks,colorlinks,citecolor=cvprblue]{hyperref}

\def\paperID{*****} % *** Enter the Paper ID here
\def\confName{CVPR}
\def\confYear{2024}

\title{Residual-based Language Models are Free Boosters for Biomedical Imaging Tasks}

\author{Zhixin Lai\thanks{These authors contributed equally to this work.} \and
        Jing Wu\protect\footnotemark[1] \and
        Suiyao Chen\protect\footnotemark[1] \and
        Yucheng Zhou \and 
        Naira Hovakimyan}

\begin{document}
\maketitle
\begin{abstract}
In this study, we uncover the unexpected efficacy of residual-based large language models (LLMs) as part of encoders for biomedical imaging tasks, a domain traditionally devoid of language or textual data. The approach diverges from established methodologies by utilizing a frozen transformer block, extracted from pre-trained LLMs, as an innovative encoder layer for the direct processing of visual tokens. This strategy represents a significant departure from the standard multi-modal vision-language frameworks, which typically hinge on language-driven prompts and inputs. We found that these LLMs could boost performance across a spectrum of biomedical imaging applications, including both 2D and 3D visual classification tasks, serving as plug-and-play boosters. More interestingly, as a byproduct, we found that the proposed framework achieved superior performance, setting new state-of-the-art results on extensive, standardized datasets in MedMNIST-2D and 3D. Through this work, we aim to open new avenues for employing LLMs in biomedical imaging and enriching the understanding of their potential in this specialized domain. Our code is available at \url{https://github.com/ZhixinLai/LLMBoostMedical}
\end{abstract}    
\section{Introduction}
\label{sec:intro}
Modern healthcare research is multifaceted, integrating various disciplines and technologies to improve patient outcomes \cite{chen2017personalized}, healthcare delivery \cite{chen2019claims}, and disease prevention \cite{ibrahim2023machine}.
One of the most critical components is biomedical imaging. The ability to classify and segment medical images accurately and swiftly is essential for clinicians, reducing errors and improving patient care. Recent advancements in artificial intelligence (AI) for vision \cite{wu2023hallucination, wu2023extended, dosovitskiy2020image, wang2023balanced}, such as Vision Transformers (ViTs), have significantly contributed to these areas. These AI models enhance the accuracy and efficiency of medical image analysis, aiding in the development of computer-aided diagnostic systems in clinical applications. By learning from large volumes of medical data, AI technologies can produce accurate diagnostic results across a range of medical applications. Their performance is often comparable to that of experienced clinicians, highlighting the transformative impact of AI in healthcare and its growing role in improving diagnostic processes.

\begin{figure}[t]
  \centering
  \includegraphics[width=1\linewidth]{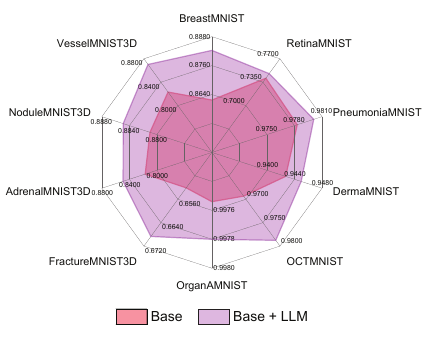}
   \caption{R-LLM benefits baseline models on a broad range of datasets in biomedical imaging tasks under the AUC metric.
   }
   \label{fig: demo}
\vspace{-5mm}
\end{figure}

Despite the promising capabilities of ViTs in biomedical imaging, we still face significant challenges that hinder further performance enhancements. First, the challenge lies in the data requirement for training these models. Effective training demands extensive, meticulously labeled datasets. Unlike other industries such as transportation \cite{chen2018data}, energy \cite{shi2017combining}, manufacturing \cite{wu2023genco, chen2017multi, chen2020optimal, bingjie2023optimal}, agriculture \cite{wu2022optimizing, tao2022optimizing}, etc., where the data collection and labeling process can be easily standardized, in the realm of biomedical imaging, creating such datasets is particularly burdensome. The need for expert knowledge is paramount due to the fine-grained nature of medical images. This process is not only time-intensive but also incurs significant financial costs, making it a substantial barrier to progress. Second, the optimization of ViT presents a critical challenge similar to the broader computer vision domain. Achieving the best performance necessitates rigorous parameter tuning, a process that requires a deep understanding of the model architecture and consumes considerable computational resources. This level of optimization, while crucial for maximizing model efficacy, is a demanding task that often stretches beyond practical limits in terms of time and computational expense. Confronted with these two significant challenges, this research focuses on exploring strategies to enhance the performance of ViT in biomedical imaging without accumulating larger datasets or dramatically increasing computational demands.

\begin{figure*}[t]
  \centering
  \includegraphics[width=0.9\linewidth]{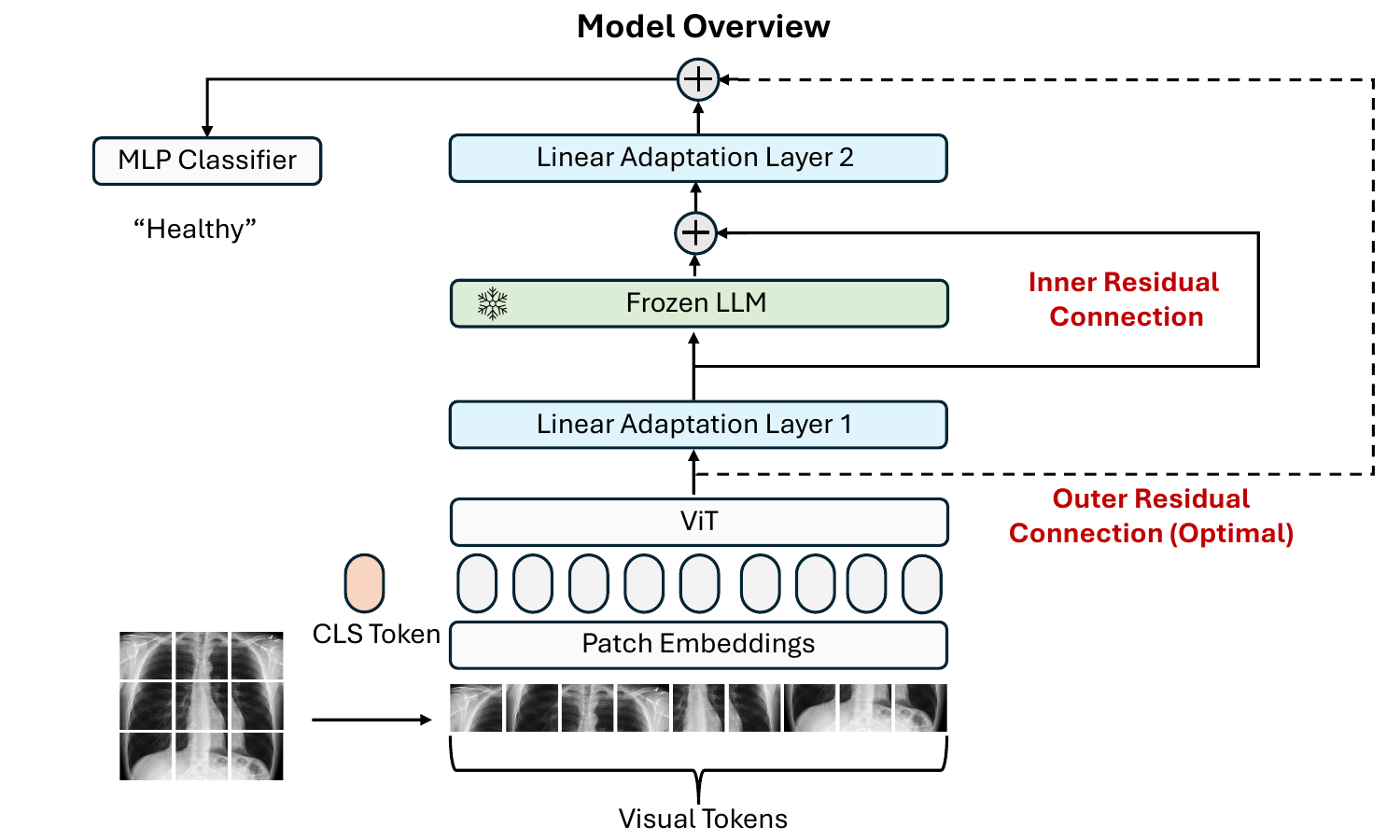}
   \caption{The proposed framework of applying language models as a booster for biomedical imaging classification task. We use Vision Transformer (ViT) from \cite{dosovitskiy2020image} for demonstration.
   }
   \label{fig:Framework}
\vspace{-5mm}
\end{figure*}

LLMs, trained on extensive textual data, have shown impressive versatility, applying their capabilities far beyond their initial linguistic applications. In computer vision, for instance, LLMs have demonstrated an intriguing capacity to engage with and interpret visual tokens, converting them into a structured, tokenized format. This integration often occurs within a multi-modal vision-language framework. Here, visual tokens are typically interfaced with LLMs through linear projection layers, or by employing cross-attention mechanisms that facilitate interaction between visual and linguistic tokens. As we delve deeper into the potential of LLMs in computer vision, a compelling question emerges: Can these models, originally designed for language processing, adeptly manage purely visual tasks, without any dependence on linguistic elements?

In pursuit of understanding the capability of LLMs in visual tasks, our research offers a novel and affirmative insight. We introduce an approach that has been largely unexplored until now: utilizing a residual-based LLM (R-LLM) block as an efficient encoder for visual data. This method is distinct in its simplicity and effectiveness, with a significant performance boost on biomedical imaging tasks, as shown in Figure~\ref{fig: demo}. Specifically, it involves three integral steps, as depicted in Figure \ref{fig:Framework}: Firstly, we integrate a frozen transformer block from an LLM into the visual encoder's architecture. Secondly, to ensure compatibility and effective information transfer, trainable linear layers are strategically positioned around the LLM block, enabling seamless feature dimension alignment. Third, a residual connection before and after the frozen LLM is introduced. Finally, while the transformer block remains frozen to retain its pre-trained characteristics, the other modules are unfrozen and undergo regular optimization during the training phase.

Remarkably, the proposed straightforward approach yields significant performance improvements across a broad range of tasks in biomedical imaging, including both 2D and 3D classification tasks. This enhancement is consistently observed with various publicly available large language models, such as LLaMA, and across different transformer blocks within these LLMs. As shown in Figure \ref{fig:Framework}-(a), the methodology innovates by treating LLM transformers as a booster of biomedical encoders, deviating significantly from the traditional perspective in vision-language models. Three key features distinguish our application of LLM transformers: First, their operation is entirely independent of language components, such as prompts, inputs, or outputs, marking a significant departure from traditional usage. Second, our method is adaptable both with and without pre-training, providing flexibility and bypassing the reliance on pre-trained models. Third, we simplify using LLMs by treating transformer blocks as distinct, modular units. This innovative approach not only challenges but also reshapes the conventional application of LLMs, particularly in the complex field of biomedical imaging tasks. In summary, our paper makes the following primary contributions:
\begin{itemize}[noitemsep,topsep=0pt]
    \item [$\bullet$] We introduce a novel residual-based framework that incorporates a frozen transformer block from pre-trained LLMs as a visual encoder layer, enhancing the learning of various biomedical imaging tasks. This innovative approach is tailored to adapt to the diverse and complex nature of biomedical images.

    \item [$\bullet$] Extensive experiments have been conducted across multiple datasets and scales, including BreastMNIST, DermaMNIST, FractureMNIST3D, etc. Surprisingly, the approach achieves state-of-the-art (SoTA) results, surpassing the performance of previous models. This underscores the effectiveness of our method in a wide array of medical imaging contexts.

    \item [$\bullet$] We provide in-depth discussions and ablation studies to dissect and understand the components of our proposed framework. These studies offer insights into the functionality and efficacy of each module, providing a comprehensive understanding of why and how our approach achieves its superior performance.

\end{itemize}
\section{Related Work}
\label{sec:related}

\subsection{Large Language Model}
In the realm of large language models, evolution began with the pretraining of transformers \cite{dosovitskiy2020image} using masked token prediction. This approach significantly enhances the versatility of language models across various tasks and modalities, which has been prominently showcased \cite{devlin2018bert, chen2024automated, chen2023you}. Following these advancements, the focus shifted towards developing larger-scale models, as guided by the scaling law \cite{kaplan2020scaling}. This direction led to the creation of groundbreaking models such as GPT \cite{brown2020language}, LLaMA \cite{touvron2023llama}, OPT \cite{zhang2022opt}, BLOOM \cite{workshop2022bloom}, and PaLM \cite{chowdhery2023palm}. These models, with their tens of billions of parameters, unveiled the potential for advanced in-context learning and exceptional zero-shot performance across various tasks, such as text classification \cite{lai2024adaptive,wang2023emp} and text infilling \cite{li2022tip}. However, the increasing complexity and size of these models presented new challenges in adaptability and efficiency. Addressing this, several papers have introduced innovative model selection \cite{hu2024validation}, transfer learning \cite{bao2023survey}, and tuning methods, such as LoRA \cite{hu2021lora} and Q-LoRA \cite{dettmers2024qlora}, which aim to enhance the flexibility of these large models without the need for extensive retraining. For our work, we build upon this foundation and unveil an interesting discovery: the transformer blocks in such LLMs possess the unique capability to interact with biomedical data.

\subsection{Vision Transformer}
The Vision Transformer introduced by \cite{dosovitskiy2020image} exemplifies how a purely transformer-based model can achieve notable success in image classification. In ViT, images are divided into patches (tokens), and transformer layers are utilized to model the global interrelations among these patches for effective classification. Building upon this, the T2T-ViT \cite{yuan2021tokens} refines the tokenization process by recursively aggregating neighboring tokens, thereby enriching the representation of local structures.  Similarly, the Swin Transformer  \cite{liu2021swin} introduces a local window-based self-attention mechanism, with a shifted window scheme for comprehensive in-window and cross-window interaction modeling. The advent of Vision Transformers (ViT) has led to an increasing number of applications \cite{zhang2023patch,yao2023improving}. In biomedical imaging, these technologies have also led to more accurate and efficient medical image segmentation and classification \cite{gao2021utnet,dai2021transmed,valanarasu2021medical}, leveraging transformers to handle variable-length inputs and capture long-distance dependencies.

\subsection{Language Models for Visual and Biomedical Imaging Tasks}
In the general vision domain, the advent of large language models (LLMs) has catalyzed a wave of innovative applications due to their generative capabilities. Notably, LLMs are being utilized to merge vision algorithms with user queries, enabling more interactive and user-specific outcomes, as explored in recent studies \cite{liu2023chatgpt,shen2024hugginggpt}. Another area of advancement is in visual programming, where LLMs play a central role in visual reasoning and in-context learning \cite{gupta2023visual,liu2023parameter}. Furthermore, the versatility of LLMs as decoders is increasingly recognized, with their ability to translate latent visual features into meaningful output tokens \cite{wang2024visionllm,zhou2024visual}. Common methodologies in this domain involve projecting visual features directly onto the input layers of LLMs  \cite{guo2023images,lin2023towards,merullo2022linearly}, or leveraging latent bottleneck structures to encode visual tokens more effectively \cite{jaegle2021perceiver,alayrac2022flamingo,li2022blip,wang2024visionllm}. 

In line with this advancement, image tasks, such as image classification \cite{ZHANG201910}, image segmentation \cite{deng2020weakly}, pattern recognition \cite{huang2024analyzing}, detection \cite{ding2019confidence}, and AR/VR technology \cite{287113}, are following this trend of using language models. Researchers in the biomedical imaging field have developed datasets that bridge the gap between vision and language \cite{irvin2019chexpert,wang2017chestx}. Utilizing these specialized datasets, significant advancements have been made in applying general-domain vision-language models to biomedical imaging \cite{boecking2022making,zhang2022contrastive,huang2021gloria}. A good example is utilizing vision-language pre-training (VLP) to incorporate domain knowledge from medicine into visual representation learning, as demonstrated in 2D and 3D image analysis \cite{liu2023t3d}. These models have shown promising results in enhancing the analysis and interpretation of medical images. However, they still require careful alignment between the visual and linguistic modalities or an additional mapping process to translate visual information into the language space.

Recent advancements in the vision domain have illuminated the potential of using transformer blocks from LLMs as general-purpose encoder layers for visual data \cite{pang2023frozen}. This perspective marks a departure from their traditional roles, primarily confined to encoding textual data, decoding tokenized outputs, or facilitating alignment between modalities. Instead, the pre-trained blocks may discern informative visual tokens and  amplify their impacts on feature representation. Inspired by this, we hypothesize that a similar idea could be effectively adapted to biomedical imaging tasks.

\section{Method}
In this section, we first introduce the overall framework of the proposed method in Section~\ref{sec: framework}. Following this, we highlight the key design and differences between the framework and previous methods in Section~\ref{sec: compare}.

\subsection{The Overall Framework}
\label{sec: framework}
We now formally introduce our comprehensive framework that harnesses the power of LLM as a free booster for biomedical imaging tasks. The entire workflow of this framework is delineated in Figure~\ref{fig:Framework}. Traditionally, the framework begins by taking a biomedical image as input, denoted as $x$. It then utilizes a vision transformer-based encoder, $\mathcal{F_{V}}$, to transform $x$ into a feature embedding $z$. This process is followed by a MLP-based classifier $\mathcal{F_{C}}$ for the final classification task, correlating with the corresponding label $y$. For the supervised learning, we define it as

\begin{equation}
\begin{aligned}
\mathcal{F_{V}}(x) &= z, \\
\mathcal{F_{C}}(z) &= y.
\end{aligned}
\end{equation}

Following the baseline framework, we incorporate a pre-trained block from LLM, specifically selecting a block from LLaMA \cite{touvron2023llama} in this study. We denote this LLM block as $\mathcal{F_{L}}$. To effectively integrate $\mathcal{F_{L}}$ into the vision-based pipeline, we introduce two additional adaptation layers: $\mathcal{F_{E}}$ and $\mathcal{F_{D}}$. The layer $\mathcal{F_{E}}$ is positioned before $\mathcal{F_{L}}$, while $\mathcal{F_{D}}$ follows it. These layers serve a critical function in aligning the dimensions between the vision data and the language model, ensuring seamless interoperability and efficient processing within our hybrid framework. Very importantly, we strategically implement a residual connection \cite{he2016deep}, positioned both before and after the LLM block. This setup allows an efficient exchange of gradient information and the passage of visual embedding through a shortcut path. Such an architecture not only facilitates the learning process but also ensures that crucial information is effectively preserved and communicated across models with different modalities, i.e., vision and language. We formally formulate this as

\begin{equation}
\begin{aligned}
\mathcal{F_{E}} \cdot \mathcal{F_{V}}(x)  &= r, \\
\mathcal{F_{D}} \cdot  (\mathcal{F_{L}}(r) + r)   &= z , \\
\mathcal{F_{C}}(z) &= y.
\end{aligned}
\end{equation}

During training, we freeze all the parameters of $\mathcal{F_{L}}$, the LLM transformer block. Meanwhile, the rest of the modules, including two adaptation layers, $\mathcal{F_{E}}$ and $\mathcal{F_{D}}$, are trained simultaneously. Following the previous paradigm \cite{pang2023frozen},  the approach modifies the behavior of LLM transformers to accommodate the stark differences between visual and textual data formats. Specifically, there are two critical adaptations. First, in LLMs, auto-regressive masks are typically used to simulate the sequential nature of text generation. However, in visual data, such as image tokens, the information is presented simultaneously rather than sequentially. Recognizing this, we forgo using auto-regressive attention masks in our framework. Instead, we employ attention masks solely to denote the presence of padded tokens in the visual data. Second, the positional embeddings utilized in LLMs, like the rotary positional embedding in LLaMA \cite{touvron2023llama}, are not typically chosen for visual encoders. Hence, for the sake of simplicity and to maintain consistency with conventional visual backbones, we opted to remove the LLMs' positional embeddings from our system.

\subsection{Comparison with Previous Methods}
\label{sec: compare}

At first glance, the proposed methods may appear akin to those used in prior vision-language model research, such as in video language retrieval \cite{lin2023towards}, FROMAGe \cite{koh2023grounding}, and LiMBeR \cite{merullo2022linearly}, where bridging the gap between vision and language spaces is achieved through linear layers. However, a distinctive aspect of our approach is the absence of an alignment between these two modalities' spaces. In essence, $\mathcal{F_{E}}$ is not constrained to map features directly from the vision to the language space, differing fundamentally from these previous methods. This conclusion and design are consistent with the previous results shown in \cite{pang2023frozen}. To be more specific, the method we propose distinguishes itself in several critical ways. Unlike prevailing approaches, it does not depend on a pre-trained encoder such as CLIP \cite{radford2021learning}, ALBEF \cite{li2021align} and Coca \cite{yu2022coca}, enabling the model to be trained entirely from scratch. This independence from pre-existing models offers greater flexibility and adaptability.
Additionally, the method functions and operates autonomously from language-based inputs or prompts, which are applicable to general biomedical imaging Tasks. Most notably, our approach represents a pioneering attempt to employ a residual connection to facilitate information exchange among different modalities, a design particularly novel in biomedical imaging. These three aspects - independence from pre-trained models, autonomy from language-based inputs, and the innovative use of residual connections across modalities - collectively underscore the distinctiveness and innovation of our method in advancing biomedical imaging technology.

\begin{table*}[t!]
    \centering
    \resizebox{1.8 \columnwidth}{!}{
    \begin{tabular}{ l  ccccccccccc cc  cc}
    \hline
    \toprule
    \textbf{Dataset}  &\multicolumn{2}{c}{\textbf{BreastMNIST}}  &\multicolumn{2}{c}{\textbf{RetinaMNIST}} &\multicolumn{2}{c}{\textbf{PneumoniaMNIST}} &\multicolumn{2}{c}{\textbf{DermaMNIST}} &\multicolumn{2}{c}{\textbf{OCTMNIST}}  &\multicolumn{2}{c}{\textbf{OrganAMNIST}} \\
    \midrule
    
   \textbf{Backbone}  &\multicolumn{2}{c}{\textbf{ViT-S}}  &\multicolumn{2}{c}{\textbf{ViT-S}} &\multicolumn{2}{c}{\textbf{ViT-S}} &\multicolumn{2}{c}{\textbf{ViT-S}} &\multicolumn{2}{c}{\textbf{ViT-S}} &\multicolumn{2}{c}{\textbf{ViT-S}}  \\

    \midrule
    \textbf{R-LLM} & \xmark & \cmark & \xmark & \cmark & \xmark & \cmark & \xmark & \cmark  & \xmark & \cmark  & \xmark & \cmark  \\
    \midrule
    
    \textbf{ACC}  & \textbf{87.17} & \textbf{87.17} & 54.25 & \textbf{57.00}  & \textbf{94.23} & 93.91 
    & 78.95 & \textbf{79.50} 
    &83.60 &\textbf{85.10}                         &95.19 &\textbf{95.22}
    
    \\

    \textbf{AUC}  & {86.17} & \textbf{88.23} & 74.09 & \textbf{74.78}  & 97.83 & \textbf{98.01} 
    & 94.27 & \textbf{94.50} 
    &96.93 &\textbf{97.88}                         &99.75 &\textbf{99.78}
    
    \\
    
    \bottomrule

    \end{tabular}
    }
    
    \caption{Performance comparison of 2D classification results of the proposed framework with and without the Residual-based LLM as a booster, evaluated using the AUC and ACC metrics. The highest-performing results are highlighted in \textbf{bold} for clarity.}
    
    \label{tab:classification2d}

\vspace{-2mm}
\end{table*}

\section{Experinments and Results}
In this section, we conduct extensive empirical evaluations and experiments to validate the effectiveness of our proposed method as a cost-free, plug-and-play booster for biomedical imaging tasks. We begin by detailing the datasets utilized in our study in Section~\ref{sec: datasets}. Subsequently, in Section~\ref{sec: cls2d}, we delve into the experiments conducted on 2D classification tasks. Following this, Section~\ref{sec: cls3d} will cover the 3D classification tasks, providing insights into the implementation details, experiments conducted, and the results derived from these tasks.
% To deepen our understanding of the proposed framework, we also present a visualization of token activation in Section~\ref{sec: visual}. 
Lastly, we conduct a series of ablation studies to understand and explore variants of the proposed method in Section~\ref{sec: ablation}.

\subsection{Datasets}
\label{sec: datasets}
We carefully selected datasets from MedMNIST V2 \cite{yang2023medmnist}, supplemented with other public datasets. Specifically, the chosen datasets encompass a broad spectrum of imaging types featuring both 2D and 3D images. Additionally, these datasets provide a diverse range of classification challenges, including both binary and multi-class tasks. 

We commence our testing with a foundational 2D dataset, comprising 780 images, to carry out binary classification tasks. This initial phase is for a preliminary evaluation of our proposed approach. Progressing from there, we expand the scale of the datasets under investigation, transitioning from hundreds to over 100,000 images. Given the limited availability of 3D datasets, our selection for 3D image analysis includes four datasets, each containing thousands of images under similar scales. We described the details of the datasets as follows:

\textbf{BreastMNIST}, drawing from a dataset of 780 breast ultrasound images \cite{al2020dataset}, classifies these images into three categories: benign, malignant, and normal. Given that the dataset comprises low-resolution images, the task has been simplified into a binary classification framework. 

\textbf{RetinaMNIST} is derived from the DeepDRiD (Deep Diabetic Retinopathy) dataset \cite{chen2021alleviating}, featuring data from 628 patients and encompassing 1600 retina fundus images. 

\textbf{PneumoniaMNIST}, adapted from an existing dataset \cite{qi2021elastic}, is comprised of 5,856 pediatric chest X-ray images. This dataset is particularly focused on the classification of pneumonia and is structured into two binary classes: `pneumonia' and `normal.' 

\textbf{DermaMNIST} is derived from the HAM10000 dataset \cite{tschandl2018ham10000}, a substantial compilation of multi-source dermatoscopic images showcasing common pigmented skin lesions. This dataset encompasses 10,015 dermatoscopic images, each with dimensions of 450 × 600 pixels. 

\textbf{OCTMNIST} is derived from a previously established dataset \cite{dataset20202nd}, consisting of 109,309 valid optical coherence tomography (OCT) images collected specifically for the study of retinal diseases. The dataset encompasses four distinct types of retinal conditions, which form the basis for a multi-class classification task. 

% \textbf{TissueMNIST}, adapted from the Broad Bioimage Benchmark Collection\cite{ljosa2012annotated}, is a dataset comprising 236,386 segmented images across eight classes of human kidney cortex cells, sourced from various reference tissue specimens.

\textbf{OrganAMNIST} originates from 3D computed tomography (CT) images utilized in the Liver Tumor Segmentation Benchmark (LiTS) \cite{acevedo2020dataset} with 58,850 images. To obtain organ labels for these images, bounding-box annotations of 11 body organs from a separate study were employed \cite{ljosa2012annotated}.

\textbf{FractureMNIST3D} is derived from the RibFrac Dataset \cite{armato2007completed}, featuring about 5,000 rib fractures from 660 CT scans. We adhere to the official dataset division for experiments.

\textbf{AdrenalMNIST3D}, derived from Zhongshan Hospital affiliated with Fudan University, encompasses shape masks from 1,584 adrenal glands (792 patients). It includes 3D shapes of adrenal glands for binary classification. This dataset is randomly divided into training, validation, and test sets, with 1,188, 98, and 298 cases, respectively, ensuring a patient-level split.

\textbf{NoduleMNIST3D}  is developed from a substantial public lung nodule dataset derived from thoracic CT scans. The dataset is partitioned in a 7:1:2 ratio into training, validation, and test sets. The images, spatially normalized to a 1mm×1mm×1mm spacing, are center-cropped to a uniform size of 28×28×28 for analysis.

\textbf{VesselMNIST3D}  comprises 103 3D brain vessel models derived from reconstructed MRA images. From these models, 1,694 healthy vessel segments and 215 aneurysm segments have been generated. The source dataset has been divided into training, validation, and test sets in a 7:1:2 ratio, facilitating a comprehensive evaluation of the models across various samples.

\begin{table*}[t]
\centering
\small % Adjust the font size
\setlength{\tabcolsep}{4pt} % Adjust spacing between columns
\begin{tabular}{ l | c   c   c   c   c   c } 
\Xhline{1pt}
\textbf{Method \textbackslash Dataset}  & \textbf{BreastMNIST} & \textbf{RetinaMNIST} & \textbf{PneumoniaMNIST} & \textbf{DermaMNIST} & \textbf{OCTMNIST} & \textbf{OrganAMNIST} \\
\hline
ResNet-18                   & 83.3    & 49.3    &86.4    &  75.4        & 76.3  &93.5

\\
ResNet-50                   & 84.2   & 51.1    &88.4    &  73.1        & 77.6  &94.7
\\
Auto-sklearn                & 80.3   & 51.5    &85.5    &  71.9        & 60.1  &76.2
\\
AutoKeras                   & 83.1   & 50.3    &87.8    &  74.9        & 76.3  &90.5
\\
Google AutoML               & 86.1   & 53.1    &94.6    &  76.8        & 77.1  &88.6
\\
MedVIT-S                    & \textbf{89.7}   & 56.1    &\textbf{96.1}    &  78.0        & 78.2  &92.8
\\
\hdashline
ViT-S + R-LLM  & 87.2   & \textbf{57.0}    &93.9    &  \textbf{79.5}        & \textbf{85.1}  &\textbf{95.2}
\\
\Xhline{1pt}
\end{tabular}
\caption{Performance comparison of 2D classification results(ACC) with the previous SoTA methods. The best values are shown in \textbf{bold}.}
\vspace{-2mm}
\label{table:sota1}
\end{table*}

\subsection{2D Classification}
\label{sec: cls2d}
We now dive into the experiments of 2D classification tasks for biomedical images. We will first introduce the detailed implementation and then move to the corresponding results. 

\subsubsection{Implementation Details}
For 2D classification experiments, all images are initially resized to a resolution of 224 x 224 pixels. We train each model using a batch size of 128, employing an AdamW optimizer for 100 epochs. The initial learning rate is set at 0.0005, coupled with a weight decay of 0.05. We utilize the ViT small model as the encoder pre-trained on ImageNet along with the llama-7b while keeping all parameters unfrozen for end-to-end training, except those in the LLaMA model. All these experiments are carried out on NVIDIA A6000 GPUs.

\subsubsection{Results}

In demonstrating the effectiveness of the R-LLM as a booster for 2D classification tasks, we primarily utilize Accuracy (ACC) and Area under the ROC Curve (AUC) as evaluation metrics. ACC, being a threshold-based metric, is particularly sensitive to class discrepancy as it evaluates discrete prediction labels. In contrast, AUC is a threshold-free metric suited for assessing continuous prediction scores. Given the diversity in dataset sizes and types in our experiments, employing both ACC and AUC provides a comprehensive assessment of our method's performance across varying conditions.

The results in Table~\ref{tab:classification2d} demonstrate that integrating the LM consistently enhances performance across various datasets and evaluation metrics. Notably, the most significant accuracy gains, approximately 1 to 3 percent, are observed in datasets such as RetinMNIST, OCTMNIST, and DermaMNIST. While improvements in other datasets are less pronounced, this could be attributed to our approach of applying a uniform set of hyperparameters across all experiments to showcase the LM's general applicability. The relatively modest enhancements in certain cases might result from this methodological choice, as it potentially limits the fine-tuning of hyperparameters tailored to each specific dataset's characteristics. Interestingly, we noticed that R-LLM did not contribute to improving the ACC metric in the PneumoniaMNIST dataset. This observation can be attributed to the dataset's imbalanced nature, with a pneumonia-to-normal ratio of approximately 3:1. Consequently, accuracy can be misleading in such an imbalanced setting, as the baseline may achieve better accuracy simply by predicting most samples as the majority class. As we switch from ACC to AUC, we can see a more fair comparison and consistently observe that R-LLM continues to benefit the classification tasks.

More surprisingly, when the LLM booster is integrated into the basic ViT model, it not only matches but, in some cases, even surpasses existing SoTA results. As outlined in Table~\ref{table:sota1}, this novel approach achieves unparalleled accuracy in datasets like BreastMNIST, RetinaMNIST, DermaMNIST, and OCTMNIST. Most notably, our method outperforms the SoTA on OCTMNIST by a remarkable margin of nearly 7 percent.

\begin{table}[t!]
    \centering
    \resizebox{1 \columnwidth}{!}{
    \begin{tabular}{ l  cccccccc  cccccccc}
    \hline
    \toprule
    \textbf{Dataset}  &\multicolumn{2}{c}{\textbf{FractureMNIST3D}}  &\multicolumn{2}{c}{\textbf{AdrenalMNIST3D}} &\multicolumn{2}{c}{\textbf{NoduleMNIST3D}} &\multicolumn{2}{c}{\textbf{VesselMNIST3D}} \\
    
    % \midrule
    % \textbf{Backbone}  &\multicolumn{2}{c}{\textbf{ViT-3D}}  &\multicolumn{2}{c}{\textbf{ViT-3D}} &\multicolumn{2}{c}{\textbf{ViT-3D}} &\multicolumn{2}{c}{\textbf{ViT-3D}} &\multicolumn{2}{c}{\textbf{ViViT}}  &\multicolumn{2}{c}{\textbf{ViViT}} &\multicolumn{2}{c}{\textbf{ViViT}} &\multicolumn{2}{c}{\textbf{ViViT}} \\
    \midrule
    
    \textbf{R-LLM} & \xmark & \cmark & \xmark & \cmark & \xmark & \cmark & \xmark & \cmark \\
    \midrule
    \textbf{ACC (ViT-3D)}  & {53.33} & \textbf{54.58} & 81.88 & \textbf{82.89}  & 86.77 & \textbf{89.68} & 90.05 & \textbf{91.10} 
     \\
    \textbf{AUC (ViT-3D)}  & {64.80} & \textbf{65.15} & 81.98 & \textbf{83.86}  & 91.48 & \textbf{92.39} & 82.55 & \textbf{83.71} 
     \\
    \midrule

    \textbf{ACC (ViViT-S)}  & {53.75} & \textbf{55.00} & 79.87 & \textbf{81.21}  & 85.81 & \textbf{86.45} & 88.74 & \textbf{90.31} 
     \\
    \textbf{AUC (ViViT-S)}  & {65.54} & \textbf{66.20} & 81.04 & \textbf{82.12}  & 86.55 & \textbf{88.76} & 83.88 & \textbf{84.56} 
     \\

    \midrule
     \textbf{ACC (ViViT-M)}  & {53.33} & \textbf{56.25} & 81.54 & \textbf{83.22}  & 85.81 & \textbf{87.42} & 89.27 & \textbf{90.58} 
     \\
    \textbf{AUC (ViViT-M)}  & {65.21} & \textbf{66.87} & 81.70 & \textbf{84.91}  & 88.10 & \textbf{88.49} & 82.31 & \textbf{87.03} 
     \\
     
    \bottomrule
    \end{tabular}
    }
    \caption{Performance comparison of 3D classification results of the proposed framework with and without the Residual-based LLM as a booster, evaluated using the AUC and ACC metrics. The highest-performing results are highlighted in \textbf{bold} for clarity.}
    \label{tab:classification3d}
\end{table}

\subsection{3D Classification}
\label{sec: cls3d}
We now move to the experiments of 3D classification tasks for biomedical images. Similarly, we will first introduce the detailed implementation and then the corresponding results.

\subsubsection{Implementation Details}
For the 3D classification experiments, each model is trained using a batch size of 128, employing an AdamW optimizer across 100 epochs. The initial learning rate is $1 \times 10^{-5}$. We adopt the ViViT \cite{arnab2021vivit} and ViT3D \cite{dosovitskiy2020image}, both modified with three channels to accommodate the 3D input, alongside the llama-7b model. The ViT3D model comprises 130.3M parameters. For ViVit, we utilize two encoder sizes: ViVit-Small (ViViT-S) and ViT-Medium (ViViT-M), containing 49.2M and 258.6M parameters, respectively. All parameters, except for those in LLaMA, are kept unfrozen for end-to-end training. These experiments are conducted on NVIDIA A6000 GPUs.

\subsubsection{Results}
Similar to the 2D datasets, we present the results for 3D datasets, reinforcing the core assertion of this paper: that LMs serve as a free booster for general bioimaging tasks, including 3D analysis. As illustrated in Table~\ref{tab:classification3d}, the results are reported for various datasets with and without the R-LLM incorporated. These results are spread across different types and scales of encoders, specifically including ViT3D, ViViT-S, and ViViT-M. Crucially, in all scenarios and across both ACC and AUC evaluation metrics, we observe marked improvements in model performance. This consistent enhancement underscores the versatility and effectiveness of the LLM as a booster in the realm of 3D biomedical imaging tasks.

For the comprehensive experiments, we follow the 2D experiment settings to compare the proposed method with previous SoTA approaches. Remarkably, in Table~\ref{table:sota2}, our framework notched three SoTA results across four datasets, without any additional hyperparameter tuning. Meanwhile, even more favorable outcomes might be attainable with further optimization and customization of training parameters.

\begin{table}[t]
\centering
\resizebox{1 \columnwidth}{!}{ % Resize table to fit the width of the text column
\begin{tabular}{ l   c c c c } 
\Xhline{1pt}
\textbf{Method \textbackslash Dataset}  & \textbf{FractureMNIST3D} & \textbf{AdrenalMNIST3D} & \textbf{NoduleMNIST3D} & \textbf{VesselMNIST3D} \\
\hline
ResNet-18 + 3D                  & 50.8    & 72.1    &84.4    &  87.7      \\
ResNet-18 + ACS             & 49.7    & 75.4    &84.7    &  \textbf{92.8}      \\
ResNet-50 + 3D                   & 49.4    & 74.5    &84.7    &  91.8      \\
ResNet-50 + ACS             & 49.4    & 78.5    &84.1    &  85.8      \\
Auto-sklearn                & 51.7    & 80.2    &87.4    &  91.5       \\
AutoKeras                   & 45.8    & 70.5    &83.4    &  89.4       \\
\hdashline
ViT3D-M + R-LLM  & {54.6}   & {82.9}    &\textbf{89.7}    &  {91.1}        \\
ViViT-M + R-LLM  & \textbf{56.3}   & \textbf{83.2}    &{87.4}    &  {90.6}        \\
\Xhline{1pt}
\end{tabular}
}
\caption{Performance(ACC) comparison of 3D classification with the previous SoTA methods. The best values are shown in \textbf{bold}.}
\label{table:sota2}
\vspace{-2mm}
\end{table}

\begin{table}[t]
\centering
\resizebox{1 \columnwidth}{!}{ % Resize table to fit the width of the text column
\begin{tabular}{ l  c c c c  } 
\Xhline{1pt}
\textbf{Method} &\textbf{Dataset} & \textbf{Num. of Parameters} & \textbf{ACC} & \textbf{AUC} \\
\hline
ViViT-M               &FractureMNIST3D &258.6M   & 53.33    & 65.21          \\
ViViT-M + MLP         &FractureMNIST3D &294.6M   & 54.17    & 65.11         \\
ViViT-M + R-LLM       &FractureMNIST3D &294.6M   & \textbf{56.25}    & \textbf{66.87}  \\

ViViT-M + R-LLM(FT)       &FractureMNIST3D &1066.62M    &53.57      &64.70

\\

\hdashline

ViViT-M               &AdrenalMNIST3D &258.4M   & 81.54    & 81.70          \\
ViViT-M + MLP         &AdrenalMNIST3D &294.6M   & 81.88    & 82.89         \\
ViViT-M + R-LLM       &AdrenalMNIST3D &294.6M   & \textbf{83.22}    & \textbf{84.91}         \\
ViViT-M + R-LLM(FT)    &AdrenalMNIST3D &1066.62M    &79.87      &81.10  

\\

\Xhline{1pt}
\end{tabular}
}
\caption{Ablation study on model capacity and fine-tuning. The best values are shown in \textbf{bold}.}
\label{table: ab1}
\end{table}

% \subsection{Visualization}
% \label{sec: visual}

\subsection{Ablation and Visualization}
\label{sec: ablation}
To further prove the effectiveness of the proposed idea and the importance of the introduced LLM block, we conduct comprehensive experiments with models of varying capacities, detailed in Section~\ref{sec: capacity}. In these experiments, we assess how the models perform with different levels of complexity. Subsequently, in Section~\ref{sec: finetune}, we explore the potential benefits of unfreezing the LLM block. This step is aimed at fully leveraging the adaptability and fitting power of the LLM. Then, we highlight the importance of residual structure in Section~\ref{sec: residual}. Lastly, Crad-CAM visualization is given in Secionn~\ref{sec: visual}.

\subsubsection{Model with Different Capacities}
\label{sec: capacity}
In evaluating the broad effectiveness of frozen LLM transformers, we considered whether the improvements could be attributed more to the expanded capacity of the linear adaptation layers, namely $\mathcal{F_{E}}$ and $\mathcal{F_{D}}$, rather than the pre-trained weights of the LLM block, $\mathcal{F_{L}}$. To investigate this, we created a variant model, ViViT-M+MLP, which has a parameter count equivalent to that of ViViT+R-LLM. This variant omits the LLM block $\mathcal{F_{L}}$, and keeps $\mathcal{F_{E}}$ and $\mathcal{F_{D}}$.

We adhered to the same training procedure outlined in Section~\ref{sec: cls3d} to ensure a fair comparison, focusing our experiments on the FractureMNIST3D and AdrenalMNIST3D datasets. The results, summarized in Table~\ref{table: ab1}, show that ViViT-M+MLP, with its increased number of parameters, does outperform the baseline ViViT-M model. However, the improvement is relatively marginal. In contrast, the enhancement observed with ViViT-M+R-LLM is both robust and substantial across both metrics. These findings lead to a significant conclusion: the pre-trained weights of the LLM transformer are instrumental to the observed improvements, and the enhancements in our biomedical imaging tasks are not merely the result of increased model capacity.

\begin{table}[t]
\centering
\resizebox{1 \columnwidth}{!}{ % Resize table to fit the width of the text column
\begin{tabular}{ l   c c c  } 
\Xhline{1pt}
\textbf{Method} &\textbf{Dataset}  & \textbf{ACC} & \textbf{AUC} \\
\hline
ViViT-M                  &FractureMNIST3D   & 53.33    & 65.21          \\
ViViT-M + R-LLM          &FractureMNIST3D    & \textbf{56.25}    & \textbf{66.87}  \\
ViViT-M + Out R-LLM      &FractureMNIST3D     & 55.83     & 65.60 \\
ViViT-M + Hybrid R-LLM   &FractureMNIST3D     & 55.00     &  65.50

\\

\hdashline

ViViT-M                  &AdrenalMNIST3D   & 81.54    & 81.70          \\
ViViT-M + R-LLM          &AdrenalMNIST3D    & \textbf{83.22}    & \textbf{84.91}  \\
ViViT-M + Out R-LLM      &AdrenalMNIST3D     & 82.55     &82.96  \\
ViViT-M + Hybrid R-LLM   &AdrenalMNIST3D     &  82.55    &  82.68

\\

\Xhline{1pt}
\end{tabular}
}
\caption{Ablation study on the importance of residual structure.}
\label{table: ab2}

\vspace{-2mm}
\end{table}

\subsubsection{End-to-end Fine-tuning}
\label{sec: finetune}

In examining whether fine-tuning the language transformer in the ViViT-M+R-LLM(FT) model is advantageous compared to maintaining it in a frozen state, we found an unexpected outcome. The results, as shown in Table~\ref{table: ab1}, indicate a decline in performance with fine-tuning, in contrast to the consistent training of the ViViT-M+R-LLM. This suggests the difficulties in training large transformer models: there is a tendency to overfit with standard-scale datasets, and fine-tuning LLMs end-to-end is often time-intensive and complex. This observation reinforces our decision to keep the LLM transformers frozen within our framework. By doing so, we simplify the training process while also ensuring effectiveness, thereby avoiding the challenges associated with fine-tuning in complex transformer architectures.

\begin{figure*}[t]
  \centering
  \includegraphics[width=0.95\linewidth]{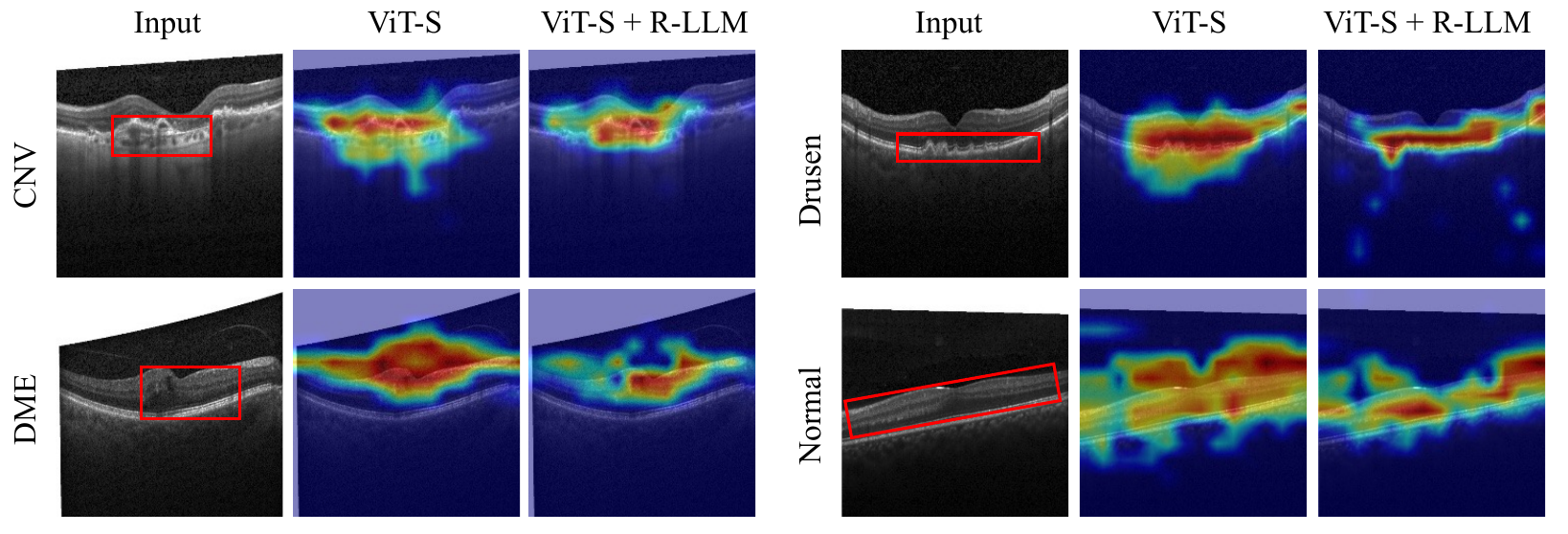}
   \caption{Visual inspection of ViT-S and ViT-S with R-LLM using Grad-CAM on original OCTMNIST dataset.}
   \label{fig:heatmap}
\vspace{-5mm}
\end{figure*}

\subsubsection{Importance of Residual Structure}
\label{sec: residual}
In this ablation study, the significance of the residual structure within our framework is meticulously examined. We found that incorporating such a structure in tandem with a Large Language Model (LLM) substantially enhances model performance. To elucidate this further, we introduce two variants of our Residual-based R-LLM: the `Out R-LLM' and the Hybrid R-LLM. Out R-LLM is designed to incorporate the residual connection before the encoder $\mathcal{F_{E}}$ and externally to the decoder $\mathcal{F_{D}}$. This can be summarized as follows:

\begin{equation}
\begin{aligned}
\mathcal{F_{V}}(x)  &= r, \\
\mathcal{F_{D}} \cdot  \mathcal{F_{L}} \cdot \mathcal{F_{E}} (r) + r   &= z , \\
\mathcal{F_{C}}(z) &= y.
\end{aligned}
\end{equation}

Hybrid R-LLM, blending the features of R-LLM and Out R-LLM, combines both internal and external residual structures. This approach offers an alternative method of integration. In line with our previous experiments, the performance of Hybrid R-LLM is evaluated on FractureMNIST3D and AdrenalMNIST3D datasets using the ACC and AUC metrics. The findings, presented in Table~\ref{table: ab2}, indicate that while R-LLM delivers the best results. However, any form of the residual structure consistently benefits the overall performance.

\subsubsection{Visual Inspection}
\label{sec: visual}

To validate the efficiency of LLM, we utilize Grad-CAM \cite{selvaraju2017grad} to qualitatively analyze the performance of ViT-S with R-LLM. We conduct training on the original OCTMNIST dataset \cite{dataset20202nd}, encompassing diverse retinal conditions: Choroidal Neovascularization (CNV), Diabetic Macular Edema (DME), Drusen, and Normal cases.

In Figure \ref{fig:heatmap}, significant regions are delineated by red rectangles, indicating areas crucial for medical diagnosis and analysis. Compared to the baseline, ViT-S enhanced with R-LLM demonstrates superior performance by closely aligning with these annotated red rectangles. This alignment enhances its ability to suppress attention toward extraneous background details effectively and to identify pivotal features essential for accurate diagnosis and analysis. This observation underscores the efficacy of our approach in medical image analysis tasks.

\section{Discussion and Conclusion}

\subsection{Discussion}
\label{sec: dicuss}

This study was primarily focused on methodically exploring a relatively under-investigated domain: the utility of pre-trained, frozen, and residual-based language transformers in biomedical imaging tasks. We have successfully demonstrated that these transformers can indeed serve as a 'free lunch', significantly boosting performance across various tasks. The experiments were carefully structured to cover a broad range of datasets and learning tasks, ensuring fair and meaningful comparisons with established baselines. Our focus was not exclusively on achieving state-of-the-art performance for every task, although this emerged as an unintended but welcome byproduct of our work.

This research not only confirms the value of LLMs in enhancing biomedical visual tasks but also opens the door for further exploration in this field. We urge fellow researchers to expand upon our work, potentially by enlarging the scope of experiments with more diverse datasets and learning tasks, not only in vision and NLP, but also Tabular \cite{sui2024table, wu2024switchtab, chen2023recontab}, Graph \cite{wei2023unleashing, chen2023graph}, etc., which could lead to more universally applicable models in the industry. Moreover, we also recognize that our approach has not yet fully harnessed the specific traits of biomedical images, such as their fine-grained structures. Delving into these aspects could yield more nuanced insights and improvements, representing a vital and promising direction for future studies.

\subsection{Conclusion}
\label{sec: conclude}
In this research, we explored the unique potential of residual-based large language models, traditionally associated with text processing, as encoders for biomedical imaging tasks. This innovative application marks a significant shift from their usual text-centric roles. By integrating a frozen transformer block from pre-trained LLMs into visual encoders as a free booster, we discovered consistent enhancements in performance across a variety of 2D and 3D biomedical imaging tasks. These findings broaden the scope of LLM applications, suggesting their utility extends well beyond language processing. Our study aims to inspire further exploration in this nascent field, particularly in bridging the modality gap between vision and language and harnessing the full potential of LLMs within the biomedical imaging domain.
{
    \small
    \bibliographystyle{ieeenat_fullname}
    \bibliography{main}
}


\end{document}